%% file: main.tex
\documentclass[10pt,twocolumn,letterpaper]{article}

\usepackage{cvpr}
\usepackage{times}
\usepackage{epsfig}
\usepackage{graphicx}
\usepackage{amsmath}
\usepackage{amssymb}
\usepackage{booktabs}
\usepackage{array}
\newcolumntype{C}[1]{>{\centering\arraybackslash}p{#1}}
\usepackage{enumitem}
\usepackage{stmaryrd}
\usepackage[toc,page]{appendix}
\input{math_commands}

\usepackage[pagebackref=true,breaklinks=true,letterpaper=true,colorlinks,bookmarks=false]{hyperref}
\usepackage{xcolor}

\hypersetup{
  colorlinks,
  breaklinks=true,
  anchorcolor={blue!50!black},
  linkcolor={blue!50!black},
  citecolor={blue!50!black},
  urlcolor={blue}
}
\cvprfinalcopy

\ifcvprfinal\fi
\begin{document}

%%%%%%%%% TITLE
\title{Semantic Bottleneck Scene Generation}

\author{\quad Samaneh Azadi$^\text{1}$\thanks{Email: sazadi@eecs.berkeley.edu},
Michael Tschannen$^\text{2}$,
Eric Tzeng$^\text{1}$,
Sylvain Gelly$^\text{2}$,
Trevor Darrell$^\text{1}$,
Mario Lucic$^\text{2}$\\[0.2cm]
$^\text{1}$University of California, Berkeley \quad $^\text{2}$Google Research, Brain Team
}

\maketitle
\thispagestyle{empty}

%%%%%%%%% ABSTRACT
\begin{abstract}
Coupling the high-fidelity generation capabilities of label-conditional image synthesis methods with the flexibility of unconditional generative models, we propose a semantic bottleneck GAN model for unconditional synthesis of complex scenes. We assume pixel-wise segmentation labels are available during training and use them to learn the scene structure. During inference, our model first synthesizes a realistic segmentation layout from scratch, then synthesizes a realistic scene conditioned on that layout. For the former, we use an unconditional progressive segmentation generation network that captures the distribution of realistic semantic scene layouts. For the latter, we use a conditional segmentation-to-image synthesis network that captures the distribution of photo-realistic images conditioned on the semantic layout. When trained end-to-end, the resulting model outperforms state-of-the-art generative models in unsupervised image synthesis on two challenging domains in terms of the Fr\'echet Inception Distance and user-study evaluations. Moreover, we demonstrate the generated segmentation maps can be used as additional training data to strongly improve recent segmentation-to-image synthesis networks.
\end{abstract}

%%%%%%%%% BODY TEXT
\section{Introduction}
Significant strides have been made on generative models for image synthesis, with a variety of methods based on Generative Adversarial Networks (GANs) \cite{GAN} achieving state-of-the-art performance.
At lower resolutions or in specialized domains, GAN-based methods are able to synthesize samples which are near-indistinguishable from real samples~\cite{biggan}. However, generating complex, high-resolution scenes from scratch remains a challenging problem.
As image resolution and complexity increase, the coherence of synthesized images decreases --- samples contain convincing local textures, but lack a consistent global structure.

Stochastic decoder-based models, such as conditional GANs, were recently proposed to alleviate some of these issues. In particular, both Pix2PixHD~\cite{pix2pixHD} and SPADE~\cite{SPADE} are able to synthesize high-quality scenes using a strong conditioning mechanism based on semantic segmentation labels during the scene generation process. Global structure encoded in the segmentation layout of the scene is what allows these models to focus primarily on generating convincing local content consistent with that structure.

A key practical drawback of such conditional models is that they require full segmentation layouts as input. Thus, unlike unconditional generative approaches which synthesize images from randomly sampled noise, these models are limited to generating images from a set of scenes that is prescribed in advance, typically either through segmentation labels from an existing dataset, or scenes that are hand-crafted by experts.
\begin{figure}[t!]
\centering
\includegraphics[width=0.9\linewidth]{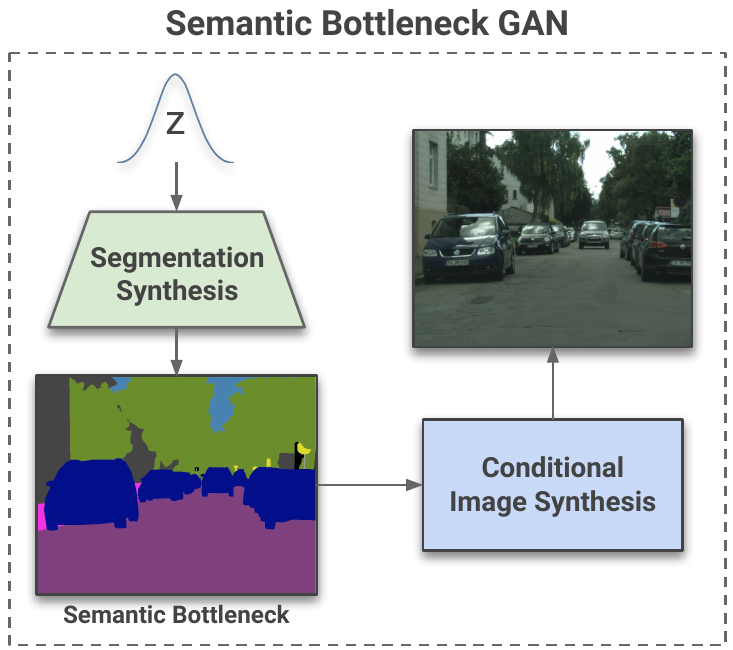}
\caption{We adversarially train the segmentation synthesis network to generate realistic segmentation maps, and then use a conditional image synthesis network to generate the final image. Fine-tuning these two components end-to-end results in state-of-the-art unconditional synthesis of complex scenes.}
\label{fig:teaser}
\end{figure}
To overcome these limitations, we propose a new model, the Semantic Bottleneck GAN, which couples high-fidelity generation capabilities of label-conditional models with the flexibility of unconditional image generation. This in turn enables our model to synthesize an unlimited number of novel complex scenes, while still maintaining high-fidelity output characteristic of image-conditional models.

Our Semantic Bottleneck GAN first unconditionally generates a pixel-wise semantic label map of a scene (i.e. for each spatial location it outputs a class label), and then generates a realistic scene image by conditioning on that semantic map. By factorizing the task into these two steps, we are able to separately tackle the problems of producing convincing segmentation layouts (i.e. a useful global structure) and filling these layouts with convincing appearances (i.e. local structure). When trained end-to-end, the model yields samples which have a coherent global structure as well as fine local details. Empirical evaluation shows that our Semantic Bottleneck GAN achieves a new state-of-the-art on two complex datasets, Cityscapes and ADE-Indoor, as measured both by the Fr\'echet Inception Distance (FID) and by user studies. Additionally, we observe that the synthesized segmentation label maps produced as part of the end-to-end image synthesis process in Semantic Bottleneck GAN can also be used to improve the performance of the state-of-the-art semantic image synthesis network~\cite{SPADE}, resulting in higher-quality outputs when conditioning on ground truth segmentation layouts. Our code will be available at \url{https://github.com/azadis/SB-GAN}.

\begin{figure*}[t!]
\centering
\includegraphics[width=\textwidth]{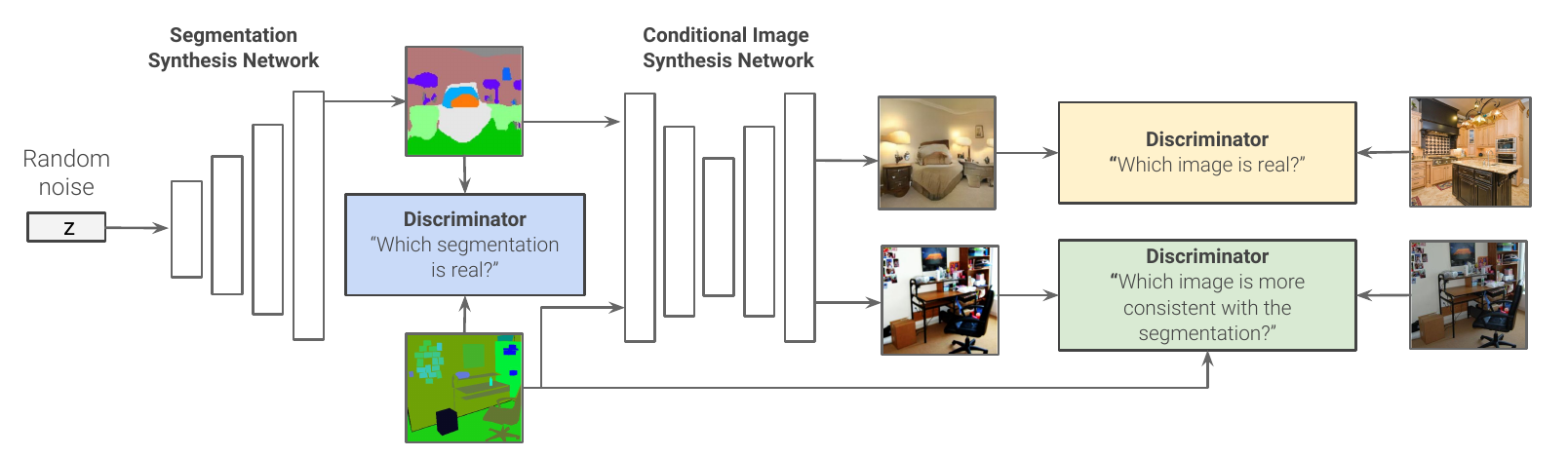}
\caption{Schematic of Semantic Bottleneck GAN. Starting from random noise, we synthesize a segmentation layout and use a discriminator to bias the segmentation synthesis network towards realistic looking segmentation layouts. The generated layout is then provided as input to a conditional image synthesis network to synthesize the final image. A second discriminator is used to bias the conditional image synthesis network towards realistic images paired with real segmentation layouts. Finally, a third unconditional discriminator is used to bias the conditional image synthesis network towards generating images that match the ground truth.}
\label{fig:end2end}
\end{figure*}

\section{Related Work}
\paragraph{Generative Adversarial Networks (GANs)}
GANs~\cite{GAN} are a powerful class of implicit generative models successfully applied to various image synthesis tasks such as image style transfer~\cite{pix2pix, cyclegan}, unsupervised representation learning~\cite{infogan, contextencoder, radford2015unsupervised}, image super-resolution~\cite{ledig2016photo, dong2016image}, and text-to-image synthesis~\cite{stackgan,attngan,qiao2019mirrorgan}. 
Training GANs is notoriously hard and recent efforts focused on improving neural architectures~\cite{PGGAN,sagan,chen2018on}, loss functions~\cite{wgan}, regularization \cite{improvedWGAN, spectralnorm}, large-scale training~\cite{biggan}, self-supervision~\cite{chen2019self}, and sampling~\cite{biggan, DRS}. One compelling approach which enables generation of high-resolution images is based on progressive training: a model is trained to first synthesize lower-resolution images (e.g. $8\times 8$), then the resolution is gradually increased until the desired resolution is achieved~\cite{PGGAN}. Recently, BigGAN~\cite{biggan} showed that GANs significantly benefit from large-scale training, both in terms of model size and batch size. We note that these models are able to synthesize high-quality images in settings where objects are very prominent and centrally placed or follow some well-defined structure, as the corresponding distribution is easier to capture. In contrast, when the scenes are more complex and the amount of data is limited, the task becomes extremely challenging for these state-of-the-art models. The aim of this work is to improve the performance in the context of complex scenes and a small number of training examples.

\paragraph{GANs on discrete domains} GANs for discrete domains have been investigated in several works~\cite{kusner2016gans,seqgan, rankgan, netgan, lu2018neural}. Training in this domain is even more challenging as the samples from discrete distributions are not differentiable with respect to the network parameters. This problem can be somewhat alleviated by using the Gumbel-softmax distribution, which is a continuous approximation to a multinomial distribution parameterized in terms of the softmax function~\cite{kusner2016gans}. We will show how to apply a similar principle to learn the distribution of discrete segmentation masks.

\vspace{-2mm}\paragraph{Conditional image synthesis} 
In conditional image synthesis one aims to generate images by conditioning on an input which can be provided in the form of an image~\cite{pix2pix, cyclegan, mcgan, compositionalGAN, liu2017unsupervised}, a text phrase~\cite{reed2016learning, stackgan, mirrorgan, ashual2019specifying, hong2018inferring}, a scene graph~\cite{johnson2018image, ashual2019specifying}, a class label or a semantic layout~\cite{odena2017conditional, chen2017photographic, pix2pixHD, SPADE}. These conditional GAN methods learn a mapping that translates samples from the source distribution into samples from the target domain. 

The text-to-image synthesis model proposed in~\cite{hong2018inferring} decomposes the synthesis task into multiple steps. First, given the text description, a semantic layout is constructed by generating object bounding boxes and refining each box by estimating object shapes. Then, an image is synthesized conditioned on the generated semantic layout from the first step. Our work shares the same high-level idea of decomposing the image generation problem into the semantic layout synthesis and the conditional semantic-layout-to-image synthesis. A key difference is that we focus on \textit{unconditional} image generation which results in a novel semantic layout generation pipeline and end-to-end network design.

\section{Semantic Bottleneck GAN (SB-GAN)}
We propose an unconditional Semantic Bottleneck GAN architecture to learn the distribution of complex scenes. To tackle the problems of learning both the global layout and the local structure, we divide this synthesis problem into two parts: an unconditional segmentation map synthesis network and a conditional segmentation-to-image synthesis model. Our first network is designed to coarsely learn the scene distribution by synthesizing semantic layouts. It generates per-pixel semantic categories following the progressive GAN model architecture (ProGAN)~\cite{PGGAN}.
The second network populates the synthesized semantic layouts with texture by predicting RGB pixel values using Spatially-Adaptive Normalization (SPADE)~\cite{SPADE}, following the architecture of the state-of-the-art semantic synthesis network in~\cite{SPADE}. We assume the ground truth segmentation masks are available for all or part of the target scene dataset. In the following sections, we will first discuss our semantic bottleneck synthesis pipeline and summarize the SPADE network for image synthesis. We will then couple these two networks in an end-to-end final design which we refer to as \textit{Semantic Bottleneck GAN (SB-GAN)}.

\subsection{Semantic bottleneck synthesis}\label{SB-network}
Our goal here is to learn a (coarse) estimate of the scene distribution from samples corresponding to real segmentation maps with $K$ semantic categories. Starting from random noise, we generate a tensor $Y \in \llbracket 1,K\rrbracket^{N \times 1 \times H \times W}$ which represents a per-pixel segmentation class, with $H$ and $W$ indicating the height and width, respectively, of the generated map and $N$ the batch size. In practice, we progressively train from a low to a high resolution using the ProGAN architecture~\cite{PGGAN} coupled with the Improved WGAN loss function~\cite{improvedWGAN} on the ground truth discrete-valued segmentation maps. In contrast to ProGAN, in which the generator outputs continuous RGB values, we predict per-pixel discrete semantic class labels. This task is extremely challenging as it requires the network to capture the intricate relationship between segmentation classes and their spatial dependencies.
To this end, we apply the Gumbel-softmax trick~\cite{jang2016categorical, maddison2016concrete} coupled with a straight-through estimator~\cite{jang2016categorical}, which we describe in detail below.

Applying a softmax function to the last layer of the generator (i.e. logits) leads to an output that can be interpreted as a probability score for each pixel belonging to each of the $K$ semantic classes. This results in probability maps $P^{ij} \in [0, 1]^{K}$, with $\sum_{k=1}^K P^{ij}_{k} =1$ for each spatial location $(i, j) \in \llbracket 1, H\rrbracket \times \llbracket 1, W \rrbracket$. To sample a semantic class from this multinomial distribution, we would ideally apply the following well-known procedure at each spatial location: (1) sample $k$ i.i.d. samples, $G_k$, from the standard Gumbel distribution, (2) add these samples to each logit, and (3) take the index of the maximal value. This reparametrization indeed allows for an efficient forward-pass, but is not differentiable. Nevertheless, the \texttt{max} operator can be replaced with the \texttt{softmax} function and the quality of the approximation can be controlled by varying the \emph{temperature hyperparameter} $\tau$---the smaller $\tau$, the closer the approximation is to the categorical distribution~\cite{jang2016categorical}:

\begin{eqnarray}
\label{gumbel}
S^{ij}_k = \frac{\exp\{(\log P^{ij}_k + G_k)/\tau\}}{\sum_{i=1}^K \exp\{(\log P^{ij}_i + G_i)/\tau\}}.
\end{eqnarray} Similar to the real samples, the synthesized samples fed to the GAN discriminator should still contain \emph{discrete} category labels. As a result, for the forward pass, we simply compute $\argmax_k S_k$, while for the backward pass, we use the soft predicted scores $S_k$ directly, a strategy also known as straight-through estimation~\cite{jang2016categorical}.

\subsection{Semantic image synthesis}
\label{SIS}
Our second sub-network converts the synthesized semantic layouts into photo-realistic images using spatially-adaptive normalization~\cite{SPADE}. The segmentation masks are employed to spread the semantic information throughout the generator by modulating the activations with a spatially adaptive learned transformation. We follow the same generator and discriminator architectures and loss functions used in~\cite{SPADE}, where the generator contains a series of SPADE residual blocks with upsampling layers. The loss functions to train SPADE are summarized as:
\begin{eqnarray}
 L_{D_{\text{SPD}}} &=& -\mathbb{E}_{y,x} [\min(0, -1 + D_{\text{SPD}}(y, x))] \nonumber\\
 && -\mathbb{E}_{y} [\min(0, -1-D_{\text{SPD}}(y,G_{\text{SPD}}(y)))] \nonumber \\
 L_{G_{\text{SPD}}} &=&  -\mathbb{E}_{y} [D_{\text{SPD}}(y, G_{\text{SPD}}(y)))] \label{eq:SPD}\\
 && + \lambda_1 L_1^{\text{VGG}} + \lambda_2 L_1^{\text{Feat}}, \nonumber
\end{eqnarray} where $G_{\text{SPD}}$, $D_{\text{SPD}}$ stand for the SPADE generator and discriminator, and $L_1^{\text{VGG}}$ and $L_1^{\text{Feat}}$ represent the VGG and discriminator feature matching $L_1$ loss functions, respectively~\cite{SPADE, pix2pixHD}. We pre-train this network using pairs of real RGB images, $x$, and their corresponding real segmentation masks, $y$, from the target scene data set.

In the next section, we will describe how to employ the synthesized segmentation masks in an end-to-end manner to improve the performance of both the semantic bottleneck and the semantic image synthesis sub-networks.
 
\subsection{End-to-end framework}
 After training semantic bottleneck synthesis model to synthesize segmentation masks and the semantic image synthesis model to stochastically map segmentations to photo-realistic images, we adversarially fine-tune the parameters of both networks in an end-to-end approach by introducing an unconditional discriminator network on top of the SPADE generator (see Figure~\ref{fig:end2end}).
 
 This second discriminator, $D_2$, has the same architecture as the SPADE discriminator, but is designed to distinguish between real RGB images and the fake ones generated from the \textit{synthesized} semantic layouts. Unlike the SPADE conditional GAN loss, which examines pairs of input segmentations and output images, $(y,x)$ in~\eqref{eq:SPD}, the GAN loss on $D_2$, $L_{D_2}$, is unconditional and only compares real images to synthesized ones, as shown in~\eqref{D2}: 
 \begin{eqnarray}
 L_{D_2} &=& -\mathbb{E}_{x} [\min(0, -1 + D_2(x))] \label{D2} \nonumber\\
 && -\mathbb{E}_{z} [\min(0, -1-D_2(G(z)))] \\
 L_G &=&  -\mathbb{E}_{z} [D_2(G(z)))] + L_{G_{\text{SPD}}} +  \lambda L_{G_{\text{SB}}} \nonumber\\
 G(z) &=& G_{\text{SPD}}(G_{\text{SB}}(z)), \nonumber
 \end{eqnarray} where $G_{\text{SB}}$ represents the semantic bottleneck synthesis generator, and $L_{G_{\text{SB}}}$ is the improved WGAN loss used to pretrain $G_{\text{SB}}$ described in Section~\ref{SB-network}.
In contrast to the conditional discriminator in SPADE, which enforces consistency between the input semantic map and the output image, $D_2$ is primarily concerned with the overall quality of the final output.
 The hyper parameter $\lambda$ determines the ratio between the two generators during fine-tuning. The parameters of both generators, $G_{\text{SB}}$ and $G_{\text{SPD}}$, as well as the corresponding discriminators, $D_{\text{SB}}$ and $D_{\text{SPD}}$, are updated in this end-to-end fine-tuning.

We illustrate our final end-to-end network in Figure~\ref{fig:end2end}.
Jointly fine-tuning the two networks in an end-to-end fashion allows the two networks to reinforce each other, leading to improved performance.
The gradients with respect to RGB images synthesized by SPADE are back-propagated to the segmentation synthesis model, thereby encouraging it to synthesize segmentation layouts that lead to higher quality final images. Hence, SPADE plays the role of a loss function for synthesizing segmentations, but in the RGB space, hence providing a goal that was absent from the initial training.
Similarly, fine-tuning SPADE with synthesized segmentations allows it to adapt to a more diverse set of scene layouts, which improves the quality of generated samples.

\section{Experiments and Results}
We evaluate the performance of the proposed approach on two datasets containing images with complex scenes, where the ground truth segmentation masks are available during training (possibly only for a subset of the images). We also study the role of the two network components, semantic bottleneck and semantic image synthesis, on the final result. We compare the performance of SB-GAN against the state-of-the-art BigGAN model~\cite{biggan} as well as a ProGAN~\cite{PGGAN} baseline that has been trained on the RGB images directly. We evaluate our method using Fr\'echet Inception Distance (FID) as well as a user study.

\begin{figure*}[t!]
\centering
\includegraphics[width=\textwidth]{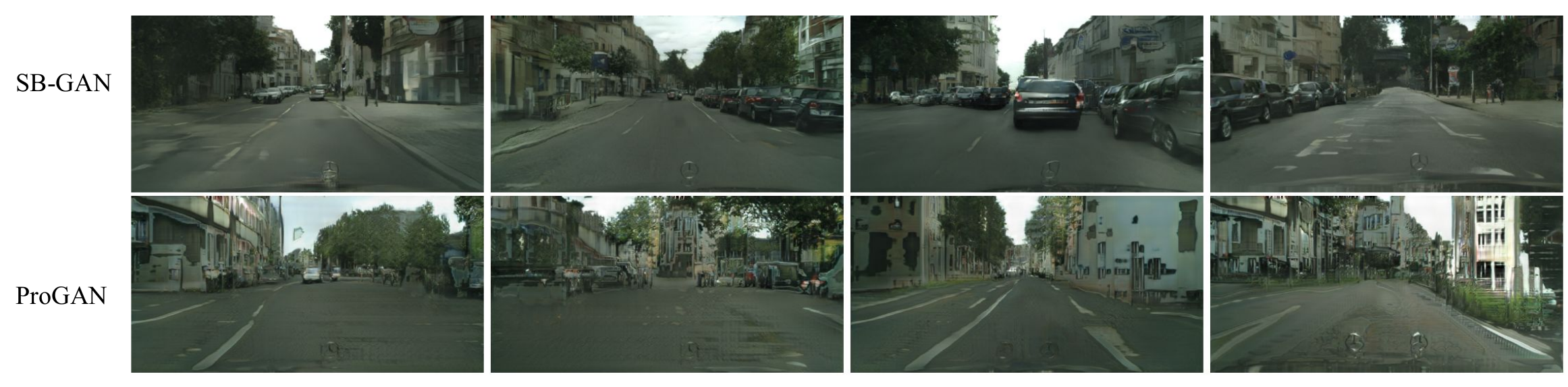}
\caption{Images synthesized by different methods trained on Cityscapes-5K. Zoom in for more detail. Although both models capture the general scene layout, SB-GAN (1st row) generates more convincing objects such as buildings and cars.}
\label{fig:city5k}
\end{figure*}

\begin{figure*}
\centering
\includegraphics[width=\textwidth]{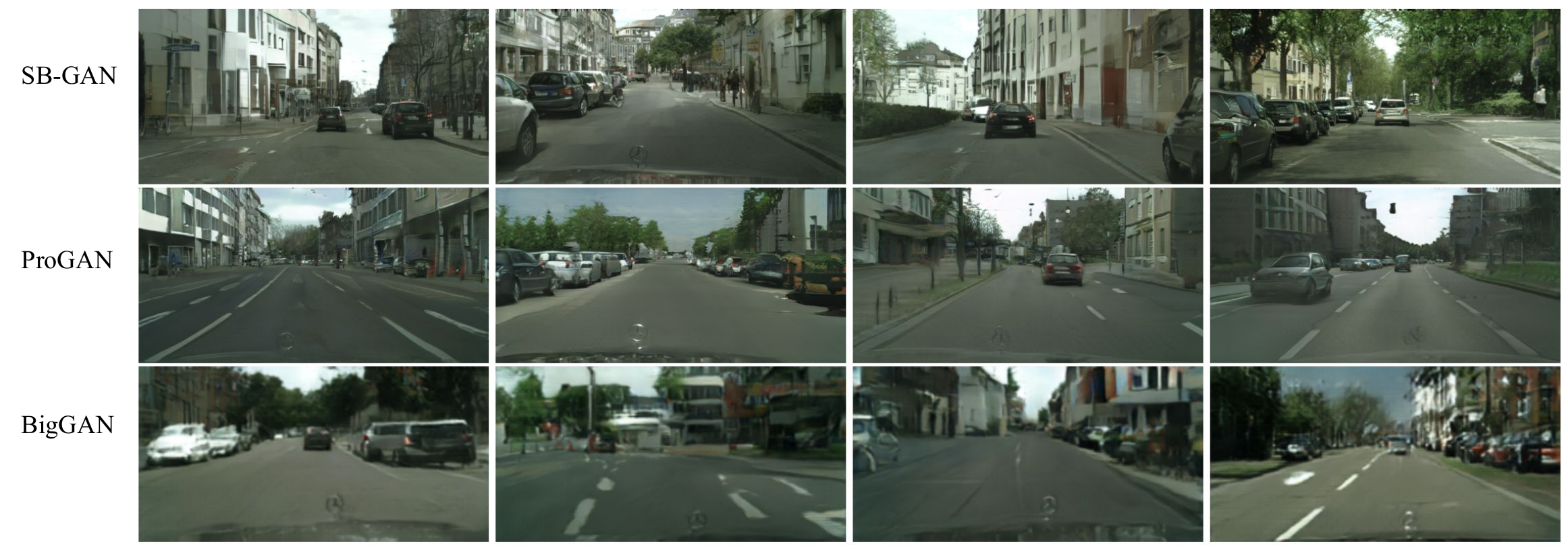}
\caption{Images synthesized by different methods trained on Cityscapes-25K. Zoom in for more detail. Images synthesized by BigGAN (3rd row) are blurry and sometimes defective in local structures.}
\label{fig:city25k}
\end{figure*}

\vspace{-2mm}
\paragraph{Datasets} We study the performance of our model on the Cityscapes and ADE-indoor datasets as the two domains with complex scene images.

\begin{itemize}[itemsep=0pt,topsep=1pt]
    \item Cityscapes-5K~\cite{cityscapes} contains street scene images in German cities with training and validation set sizes of 3,000 and 500 images, respectively. Ground truth segmentation masks with 33 semantic classes are available for all images in this dataset.
    \item Cityscapes-25K~\cite{cityscapes} contains street scene images in German cities with training and validation set sizes of 23,000 and 500 images, respectively. Cityscapes-5K is a subset of this dataset, providing 3,000 images in the training set here as well as the entire validation set. Fine ground truth annotations are only provided for this subset, with the remaining 20,000 training images containing only coarse annotations.
    We extract the corresponding fine annotations for the rest of training images using the state-of-the-art segmentation model~\cite{DRN, Yu2016} trained on the training annotated samples from Cityscapes-5K. This dataset contains 19 semantic classes. 
    \item ADE-Indoor is a subset of the ADE20K dataset~\cite{ADE20k} containing 4,377 challenging training images from indoor scenes and 433 validation images with 95 semantic categories. 
\end{itemize}

\begin{figure*}
\centering
\includegraphics[width=\textwidth]{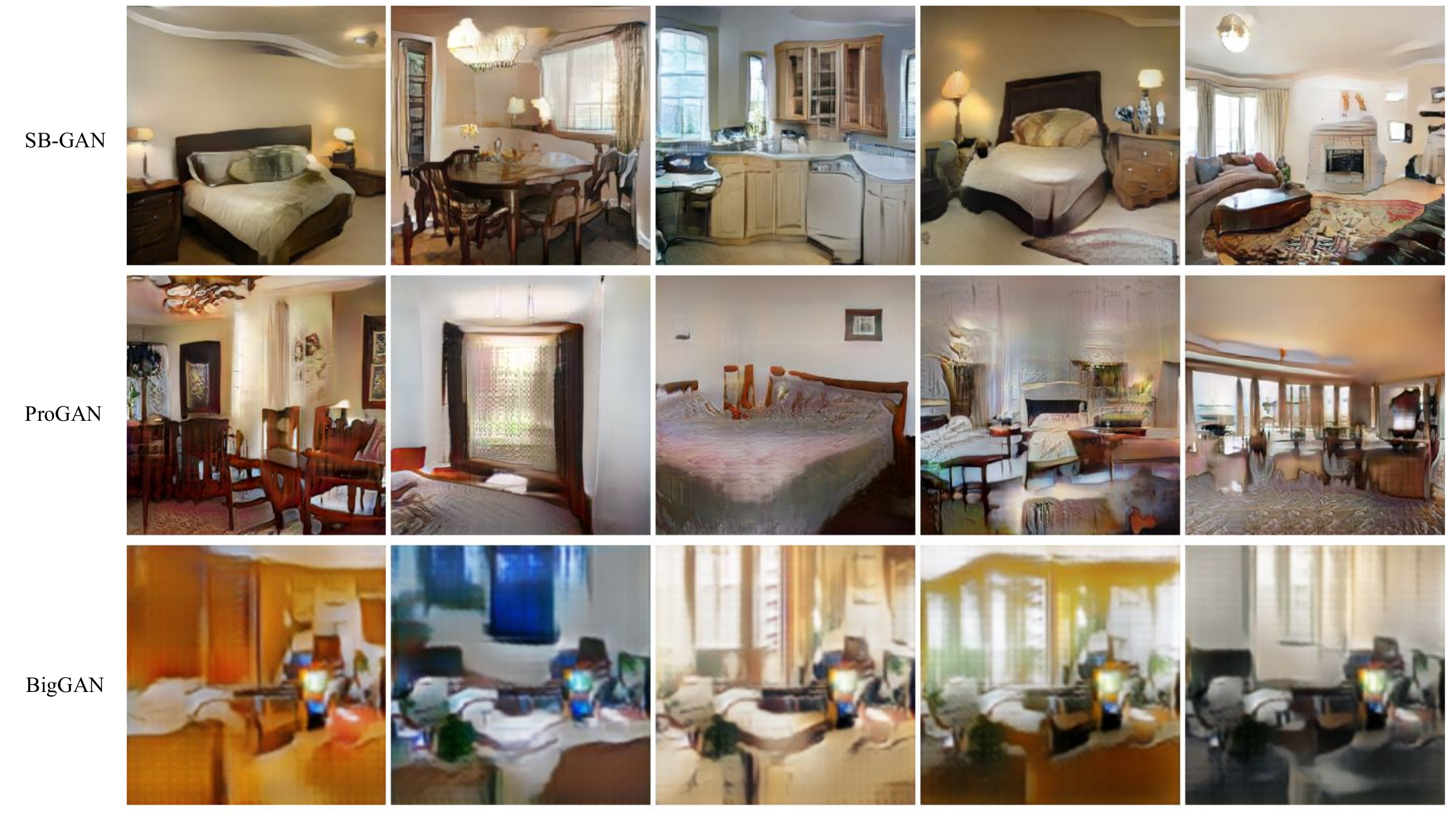}
\caption{Images synthesized by different methods trained on ADE-Indoor. This dataset is very challenging, causing mode collapse for the BigGAN model (3rd row). In contrast, samples generated by SB-GAN (1st row) are generally of higher quality and much more structured than those of ProGAN (2nd row).}
\label{fig:ade}
\end{figure*}

\vspace{-3mm}
\paragraph{Evaluation} We use the Fr{\'e}chet Inception Distance (FID)~\cite{heusel2017gans} as well as a user study to evaluate the quality of the generated samples. To compute FID, the real data and generated samples are first embedded in a specific layer of a pre-trained Inception network. Then, a multivariate Gaussian is fit to the data, and the distance is computed as
$\FID(x, g) = ||\mu_x -\mu_g||_2^2 + \Tr(\Sigma_x + \Sigma_g - 2(\Sigma_x\Sigma_g)^\frac12)$,
where $\mu$ and $\Sigma$ denote the empirical mean and covariance, and subscripts $x$ and $g$ denote the real and generated data respectively. FID is shown to be sensitive to both the addition of spurious modes and to mode dropping~\cite{sajjadi2018assessing,lucic2018}. On the Cityscapes dataset, we ran five trials where we computed FID on 500 random synthetic images and 500 real validation images, and report the average score. On ADE-Indoor, the same process is repeated on batches of 433 images.  

\paragraph{Implementation details} In all our experiments, we set $\lambda_1 = \lambda_2 = 10$, and $\lambda = 10$. The initial generator and discriminator learning rates for training SPADE both in the pretraining and end-to-end steps are $10^{-4}$ and $4\cdot10^{-4}$, respectively. The learning rate for the semantic bottleneck synthesis sub-network is set to $10^{-3}$ in the pretraining step and to $10^{-5}$ in the end-to-end fine-tuning on Cityscapes, and to $10^{-4}$ for ADE-Indoor. The temperature hyperparameter, $\tau$, is always set to 1.
For BigGAN, we followed the setup in~\cite{luvcic2019high}\footnote{Configuration as in \url{https://github.com/google/compare_gan/blob/master/example_configs/biggan_imagenet128.gin}}, where we modified the code to allow for non-square images of Cityscapes. We used one class label for all images to have an unconditional BigGAN model. For both datasets we varied the batch size (using values in $\{128, 256, 512, 2048\}$), the learning rate, and the location of the self-attention block. We trained the final model for $50$K iterations.

\subsection{Qualitative results}
In Figures~\ref{fig:city5k},~\ref{fig:city25k}, and~\ref{fig:ade}, we provide qualitative comparisons of the competing methods on the three aforementioned datasets. We observe that both Cityscapes-5K and ADE-Indoor are very challenging for the state-of-the-art ProGAN and BigGAN models, likely due to the complexity of the data and small number of training instances. Even at a resolution of $128\times 128$ on the ADE-Indoor dataset, BigGAN suffers from mode collapse, as illustrated in the last row of Figure~\ref{fig:ade}. In contrast, SB-GAN significantly improves on the structure of the scene distribution, and provides samples of higher quality. On Cityscapes-25K, the performance improvement of SB-GAN is more modest due to the large number of training images available. It is worth emphasizing that in this case only 3K ground truth segmentations for training SB-GAN are available. Compared to BigGAN, images synthesized by SB-GAN are sharper and contain more structural details (e.g., one can zoom-in on the synthesized cars). More qualitative examples are presented in the Appendix.

\subsection{Quantitative evaluation}
To provide a thorough empirical evaluation of the proposed approach, we generate samples for each dataset and report the FID scores of the resulting images (averaged across 5 sets of generated samples).
We evaluate SB-GAN both before and after end-to-end fine-tuning, and compare our method to two strong baselines, ProGAN~\cite{PGGAN} and BigGAN~\cite{biggan}. The results are detailed in Tables~\ref{table:results} and~\ref{table:results128}.

\begin{table}[h]
\setlength{\tabcolsep}{3pt}
\setlength{\extrarowheight}{5pt}
\renewcommand{\arraystretch}{0.75}
\centering
\begin{tabular}{l C{1.2cm}  C{1.7cm}C{1.8cm}}
\toprule
& \multicolumn{3}{c}{\textsc{Method}} \\ \cmidrule{2-4}
               & ProGAN  & SB-GAN W/O FT & SB-GAN \\ \midrule
\textsc{Cityscapes-5k}  &  92.57   & 83.20       & \textbf{65.49} \\
\textsc{Cityscapes-25k} &  63.87   & 71.13       & \textbf{62.97} \\
\textsc{ADE-Indoor}     & 104.83  & 91.80       &  \textbf{85.27}\\
\bottomrule
\end{tabular}
\vspace{2mm}
\caption{FID of the synthesized samples (lower is better), averaged over 5 random sets of samples. Images were synthesized at resolution of $256\times 512$ on Cityscapes and $256\times 256$ on ADE-Indoor. }
\label{table:results}
\end{table}

\begin{figure*}[h]
\centering
\includegraphics[width=\textwidth]{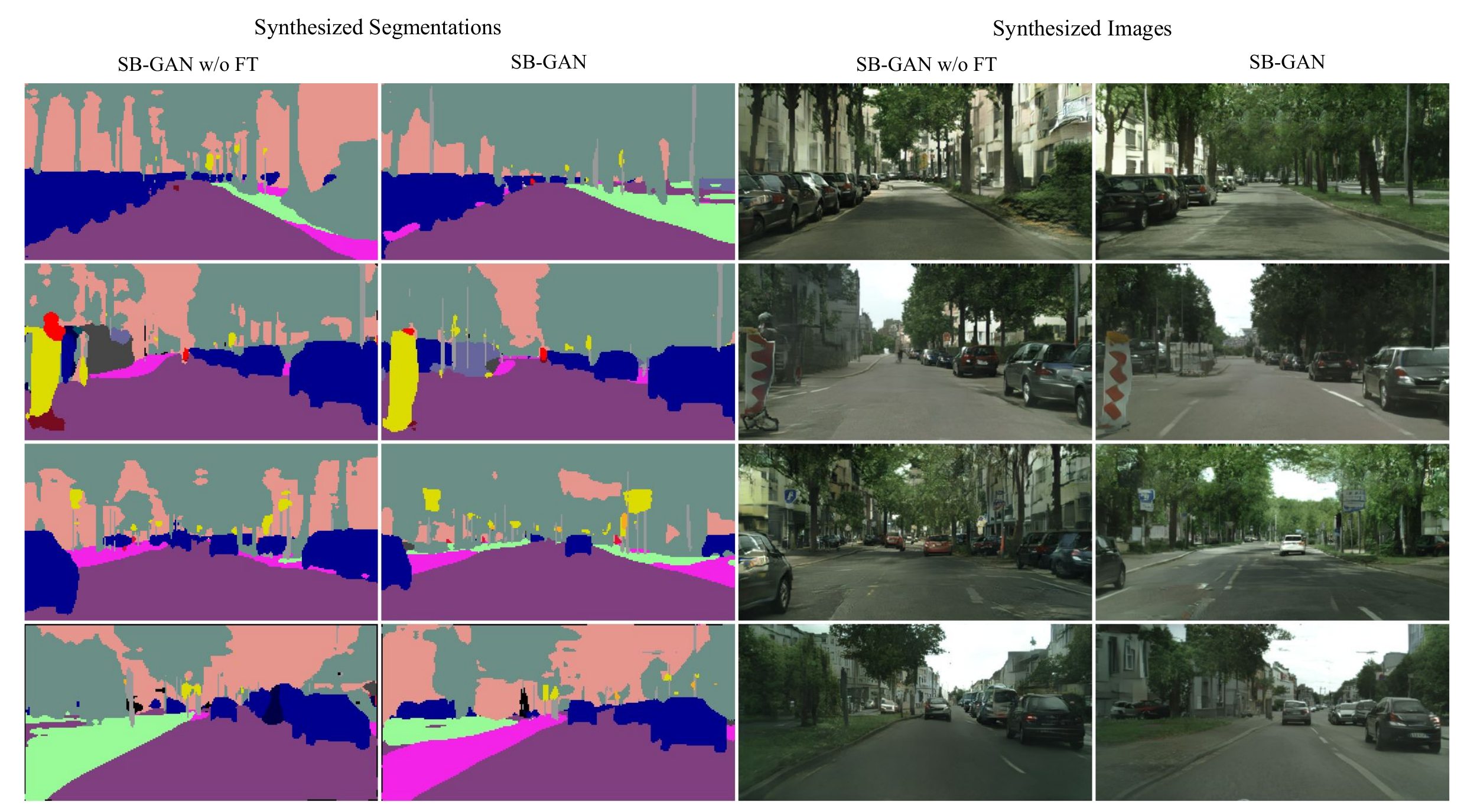}
\caption{The effect of fine-tuning on the baseline setup for the Cityscapes-25K dataset. We observe that both the global structure of the segmentations and the performance of semantic image synthesis improve after fine-tuning, resulting in images of higher quality. \label{fig:ablate_city25k}}
\end{figure*}

 First, in the low-data regime, even without fine-tuning, our Semantic Bottleneck GAN produces higher quality samples and significantly outperforms the baselines on Cityscapes-5K and ADE-Indoor. The advantage of our proposed method is even more striking on smaller datasets. While competing methods are unable to learn a high-quality model of the underlying distribution without having access to a large number of samples, SB-GAN is less sensitive to the number of training data points. Secondly, we observe that by jointly training the semantic bottleneck and image synthesis components, SB-GAN produces state-of-the-art results across all three datasets.

We were not able to successfully train BigGAN at a resolution of $256\times 512$ due to instability observed during training and mode collapse. Table~\ref{table:results128}, however, shows the results for a lower-resolution setting, for which we were able to successfully train BigGAN. We report the results before the training collapses. BigGAN is, to a certain extent, able to capture the distribution of Cityscapes-25K, but fails completely on ADE-Indoor. Interestingly, BigGAN fails to capture the distribution of Cityscapes-5K even at $128\times128$ resolution. The standard deviation of the FID scores computed in Tables~\ref{table:results} and~\ref{table:results128} is within $1.5\%$ of the mean for Cityscapes and within $3\%$ of the mean for ADE-Indoor.

\begin{table}[h!]
\setlength{\tabcolsep}{4pt}
\setlength{\extrarowheight}{5pt}
\renewcommand{\arraystretch}{0.75}
\centering
\begin{tabular}{l C{1.2cm} C{1.5cm}C{1.5cm}}
\toprule
 & \multicolumn{3}{c}{\textsc{Method}} \\ \cmidrule{2-4}
               & ProGAN  & BigGAN & SB-GAN \\ \midrule
\textsc{Cityscapes-25k} &  56.7   &  64.82   & \textbf{54.92} \\
\textsc{ADE-Indoor}      & 85.94 & 156.65  &  \textbf{81.39}\\
\bottomrule
\end{tabular}
\vspace{2mm}
\caption{FID of the synthesized samples (lower is better), averaged over 5 random sets of samples. Images were synthesized at resolution of $128\times 256$ on Cityscapes and $128\times 128$ on ADE-Indoor.}
\label{table:results128}
\end{table}

\begin{figure*}[t!]
\centering
\includegraphics[width=\textwidth]{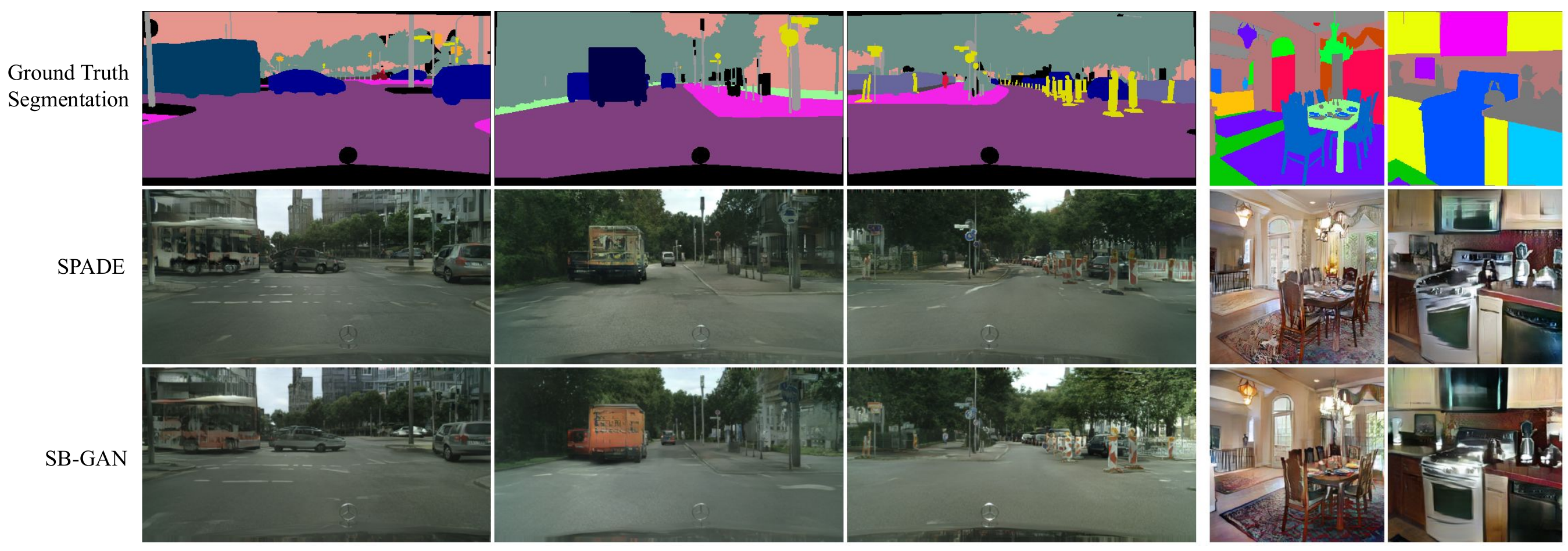}
\caption{The effect of SB-GAN on improving the performance of the state-of-the-art semantic image synthesis model (SPADE~\cite{SPADE}) on ground truth segmentations of Cityscapes-25K (left) and ADE-Indoor (right) validation sets. For SB-GAN, we train the entire model end-to-end, extract the trained SPADE sub-network, and synthesize samples conditioned on the ground truth labels.}
\label{fig:ablate_spade}
\end{figure*}

\paragraph{Generating by conditioning on real segmentations}
To independently assess the impact of end-to-end training on the conditional image synthesis sub-network, we evaluate the quality of generated samples when conditioning on ground truth validation segmentations from each dataset. 
Comparisons to the baseline network SPADE~\cite{SPADE} are provided in Table~\ref{table:gtlabel-results} and Figure~\ref{fig:ablate_spade}. We observe that the image synthesis component of SB-GAN consistently outperforms SPADE across all three datasets, indicating that fine-tuning on synthetic labels produced by the segmentation generator improves the conditional image generator. Please refer to the Appendix for more qualitative examples.

\begin{table}[h]
\setlength{\tabcolsep}{4pt}
\setlength{\extrarowheight}{5pt}
\renewcommand{\arraystretch}{0.75}
\centering
\begin{tabular}{lcc}
\toprule
  & \multicolumn{2}{c}{\textsc{Method}} \\ \cmidrule{2-3}
& SPADE & SB-GAN \\ \midrule
\textsc{Cityscapes-5k}     & 72.12 & \textbf{60.39} \\
\textsc{Cityscapes-25k}    & 60.83 & \textbf{54.13} \\
\textsc{ADE-Indoor}        & 50.30 & \textbf{48.15} \\
\bottomrule
\end{tabular}
\vspace{3mm}
\caption{FID of the synthesized samples when conditioned on the ground truth labels (lower is better), averaged over 5 random sets of samples. For SB-GAN, we train the entire model end-to-end, extract the trained SPADE sub-network, and synthesize samples conditioned on the ground truth labels.}
\label{table:gtlabel-results}
\end{table}

\begin{figure*}[t!]
\centering
\includegraphics[width=\textwidth]{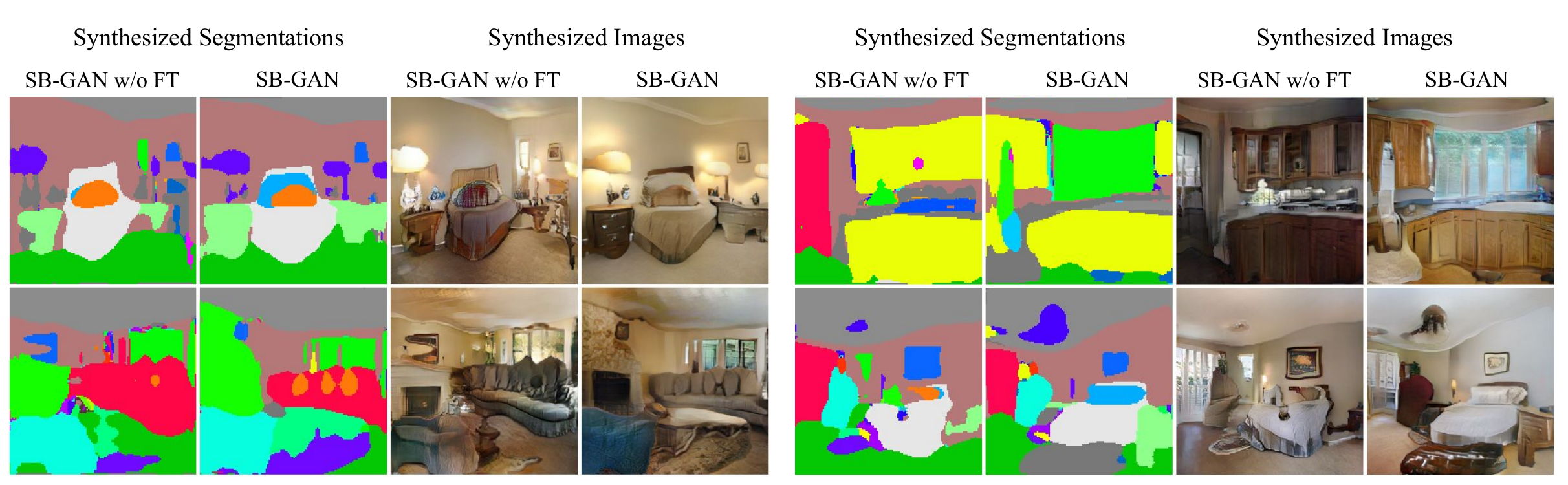}
\caption{The effect of fine-tuning (FT) on the baseline setup for ADE-Indoor dataset. We observe that both the global structure of the segmentations and the performance of semantic image synthesis have been improved after fine-tuning, resulting in images of higher quality.}
\label{fig:ablate_ade}
\end{figure*}

\paragraph{Fine-tuning ablation study}
To further dissect the effect of end-to-end training, we perform a study 
on different components of SB-GAN. In particular, we consider three settings: (1) SB-GAN before end-to-end fine-tuning, (2) fine-tuning only the semantic bottleneck synthesis component, (3) fine-tuning only the conditional image synthesis component, and (4) fine-tuning all components jointly. The results on the Cityscapes-5K dataset (resolution $128\times 256$) are reported in Table~\ref{table:ablation}. Finally, the impact of fine-tuning on the quality of samples can be observed in Figures~\ref{fig:ablate_city25k} and ~\ref{fig:ablate_ade}. 

\begin{table}[h]
\setlength{\tabcolsep}{4pt}
\setlength{\extrarowheight}{5pt}
\renewcommand{\arraystretch}{0.75}
\centering
\begin{tabular}{C{1.5cm} C{2cm} C{2cm}C{1.3cm}}
\toprule
\multicolumn{4}{c}{\textsc{Method}} \\ \cmidrule{1-4}
    No FT    & FT SB    & FT SPADE & FT Both \\ \midrule
  70.15 & 66.22   &  63.04  & \textbf{58.67} \\
\bottomrule
\end{tabular}
\vspace{2mm}
\caption{Ablation study of various components of SB-GAN. We report FID scores of SB-GAN before fine-tuning, fine-tuning only the semantic bottleneck synthesis component, fine-tuning only the image synthesis component, and full end-to-end fine-tuning. Experiments are performed on the Cityscapes-5K dataset at a resolution of $128\times 256$.}
\label{table:ablation}
\vspace{-2mm}
\end{table}

\subsection{Human evaluation}
We used Amazon Mechanical Turk (AMT) to study and compare the performance of different methods in terms of user assessments. We evaluate the performance of each model on each dataset through $\sim$600 pairs of (synthesized images, evaluators) containing 200 unique synthesized images. For each image, evaluators were asked to select a quality score from 1 to 4, indicating terrible and high quality images, respectively. Results are summarized in Table~\ref{table:userstudy} and are consistent with FID-based evaluations, with SB-GAN as the winner in all datasets once again.

\begin{table}[h]
\setlength{\tabcolsep}{5pt}
\setlength{\extrarowheight}{5pt}
\renewcommand{\arraystretch}{0.75}
\centering
\begin{tabular}{lccc}
\toprule
& \multicolumn{3}{c}{\textsc{Method}} \\ \cmidrule{2-4}
& ProGAN & BigGAN & SB-GAN \\ \midrule
\textsc{Cityscapes-5k}     & 2.08 & - & \textbf{2.48} \\
\textsc{Cityscapes-25k}    & 2.53 & 2.27 & \textbf{2.61} \\
\textsc{Ade-Indoor}        & 2.35 & 1.96 & \textbf{2.49} \\
\bottomrule
\end{tabular}
\vspace{2mm}
\caption{Average user evaluation scores when each user has selected a quality score in the range of 1 (terrible quality) to 4 (high quality) for each image.}
\label{table:userstudy}
\vspace{-2mm}
\end{table}

\vspace{-1mm}
\section{Conclusion}
We proposed an end-to-end Semantic Bottleneck GAN model that synthesizes semantic layouts from scratch, and then generates photo-realistic scenes conditioned on the synthesized layouts. Through extensive quantitative and qualitative evaluations, we showed that this novel end-to-end training pipeline significantly outperforms the state-of-the-art models in unconditional synthesis of complex scenes. In addition, Semantic Bottleneck GAN strongly improves the performance of the state-of-the-art semantic image synthesis model in synthesizing photo-realistic images from ground truth segmentations.

We believe that the idea of applying a semantic bottleneck to other generative models should be explored in future work. In addition, novel ways to train GANs with discrete outputs could be explored, especially techniques to deal with the non-differentiable nature of the generated outputs.

\paragraph{Acknowledgments}
This work was supported by Google through Google Cloud Platform research credits. We thank Marvin Ritter for help with issues related to the compare\_gan library \cite{lucic2018}. We are grateful to the members of BAIR for fruitful discussions. SA is supported by the Facebook graduate fellowship. 
%------------------------------------------------------------------------

%\newpage
{\small
\bibliographystyle{ieee_fullname}
\bibliography{egbib}
}

\newpage
\appendix
\section*{Appendix}
\section{Additional results}
\begin{figure*}[t!]
\centering
\includegraphics[width=\textwidth]{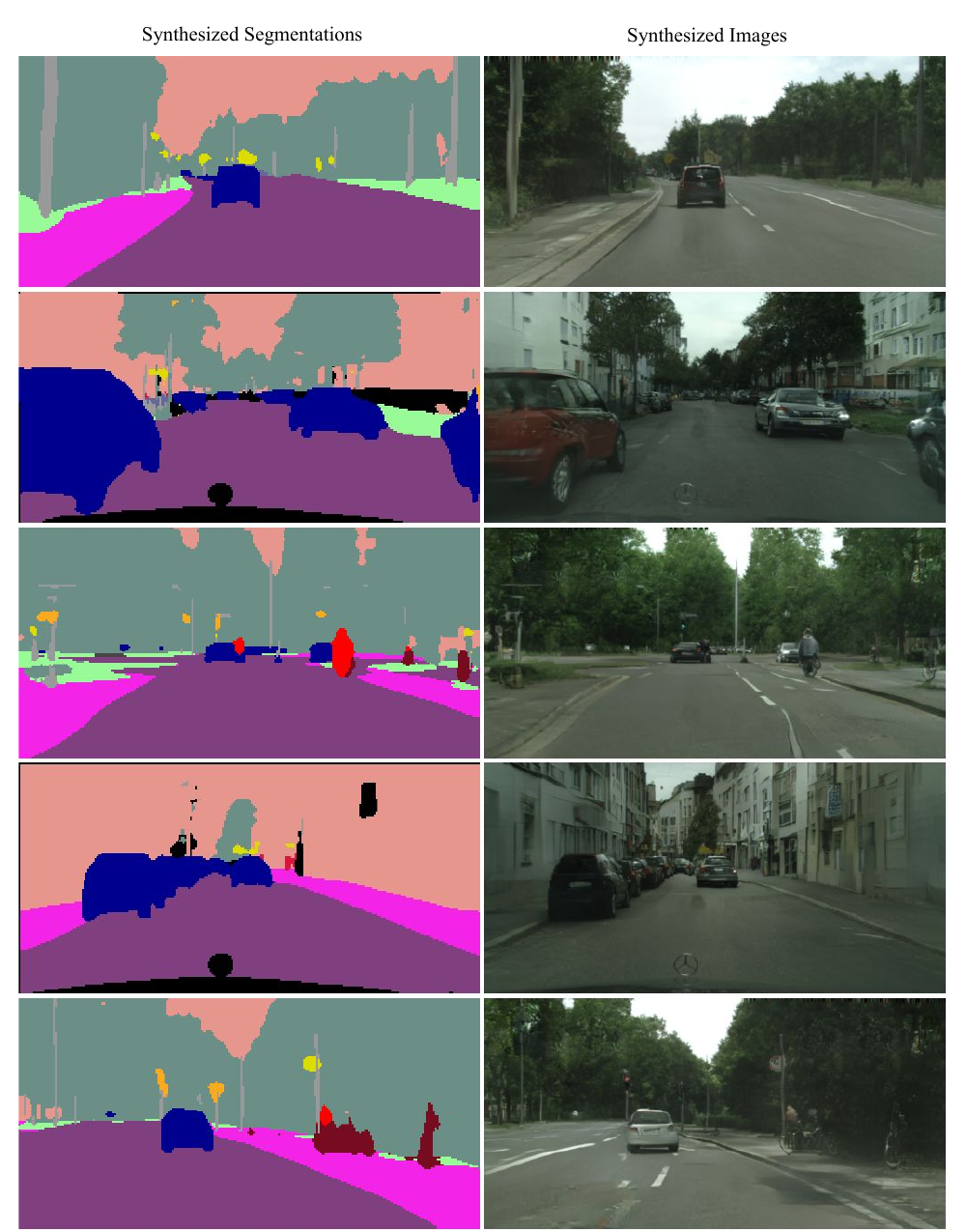}
\caption{Segmentations and their corresponding images synthesized by SB-GAN trained on the Cityscapes-25K dataset.}
\label{fig:ablate_city}
\end{figure*}

\begin{figure*}[t!]
\centering
\includegraphics[width=\textwidth]{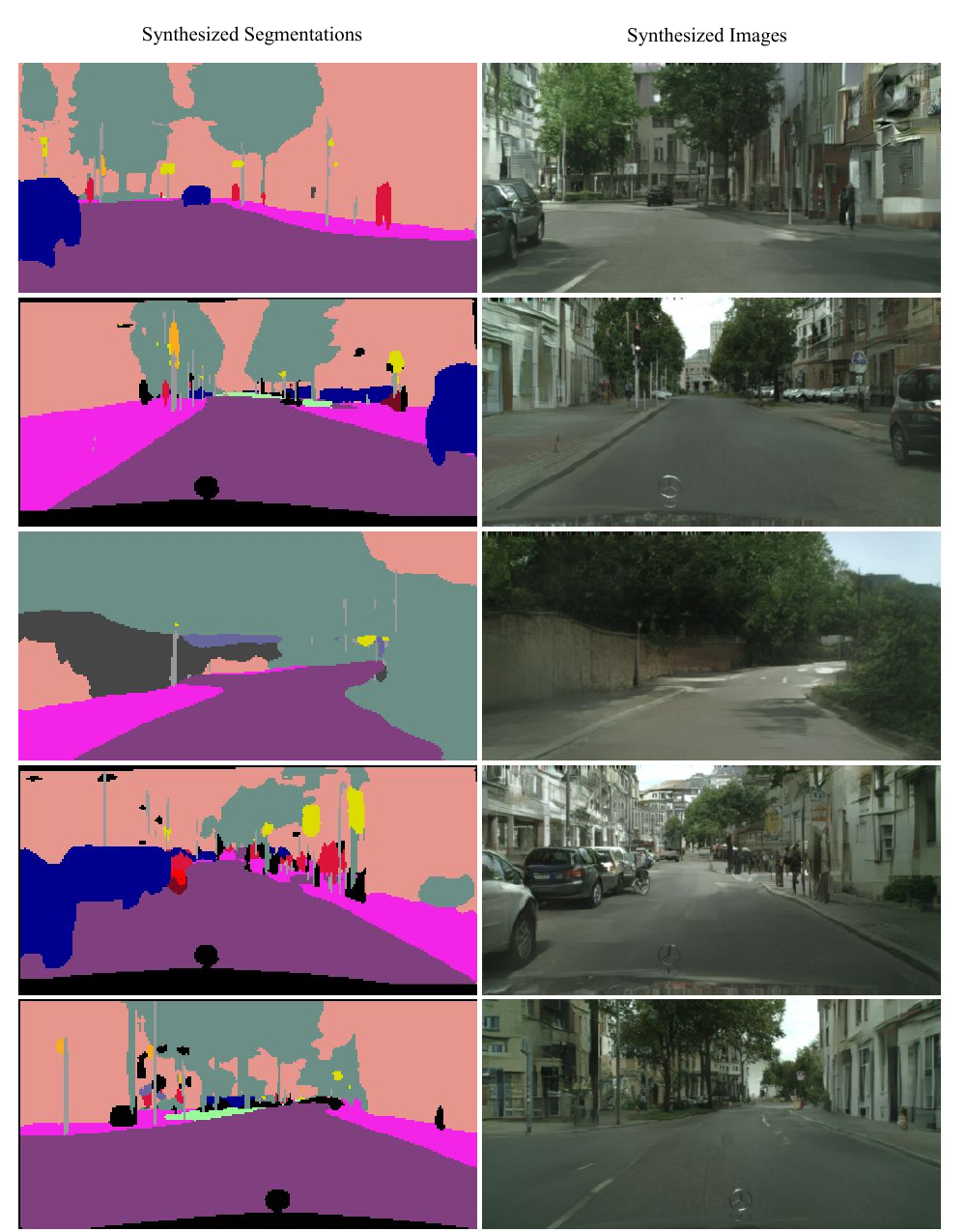}
\caption{Segmentations and their corresponding images synthesized by SB-GAN trained on the Cityscapes-25K dataset.}
\label{fig:ablate_city2}
\end{figure*}

\begin{figure*}[t!]
\centering
\includegraphics[width=\textwidth]{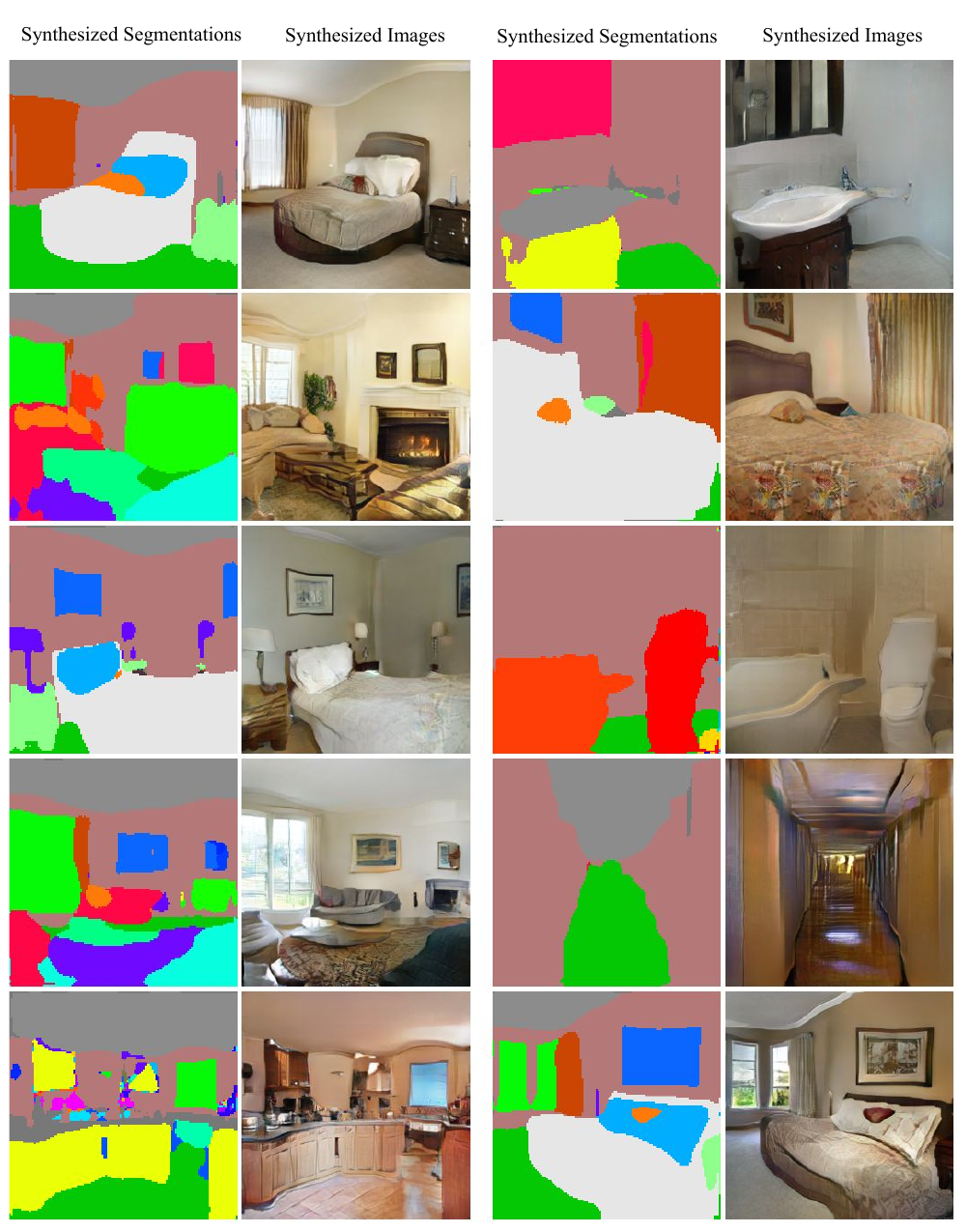}
\caption{Segmentations and their corresponding images synthesized by SB-GAN trained on the ADE-Indoor dataset.}
\label{fig:ablate_ade2}
\end{figure*}

In Figures~\ref{fig:ablate_city},~\ref{fig:ablate_city2},~\ref{fig:ablate_ade}, and~\ref{fig:ablate_ade2}, we show additional synthetic results from our proposed SB-GAN model including both the synthesized segmentations and their corresponding synthesized images from the Cityscapes-25K and ADE-Indoor datasets. As mentioned in the paper, on the Cityscapes-25K dataset, fine ground truth annotations are only provided for the Cityscapes-5k subset. We extract the corresponding fine annotations for the rest of training images using the state-of-the-art segmentation model~\cite{DRN, Yu2016} trained on the training annotated samples from Cityscapes-5K. 

\begin{figure*}[!t]
\centering
\includegraphics[width=\textwidth]{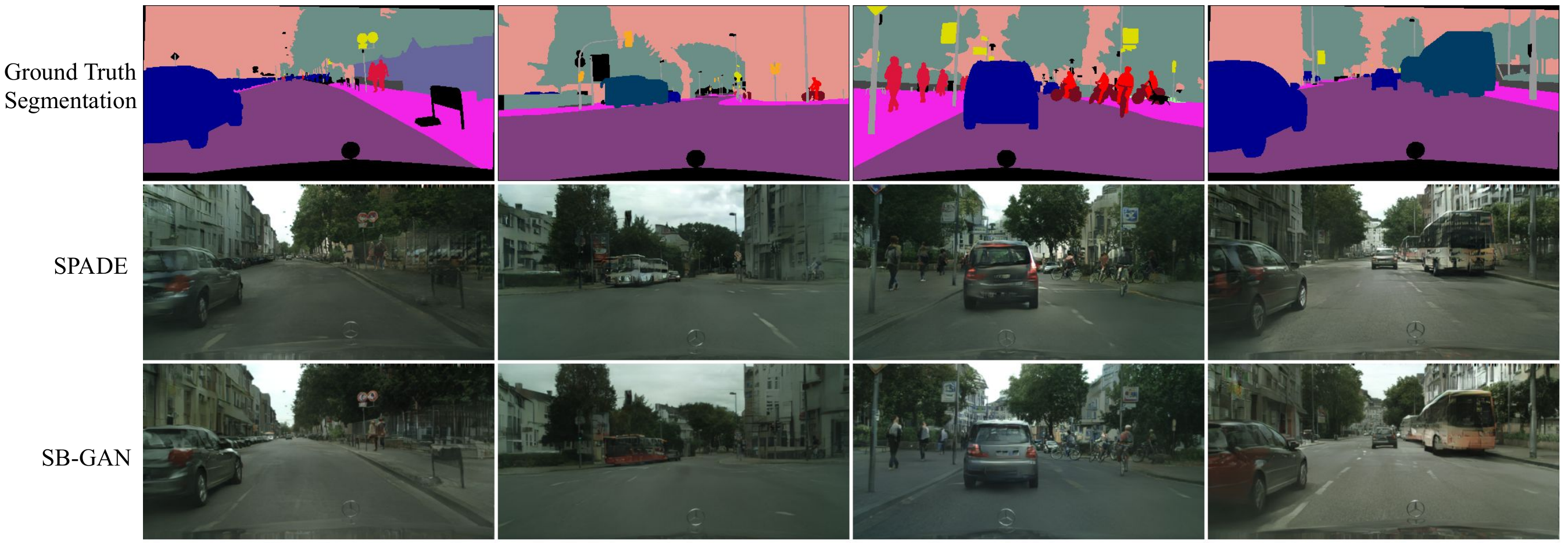}
\caption{The effect of SB-GAN on improving the performance of the state-of-the-art semantic image synthesis model (SPADE) on ground truth segmentations of Cityscapes-25K validation set. For SB-GAN, we train the entire model end-to-end, extract the trained SPADE sub-network, and synthesize samples conditioned on the ground truth labels.}
\label{fig:ablate_city_spade}
\end{figure*}

\begin{figure*}[t!]
\centering
\includegraphics[width=\textwidth]{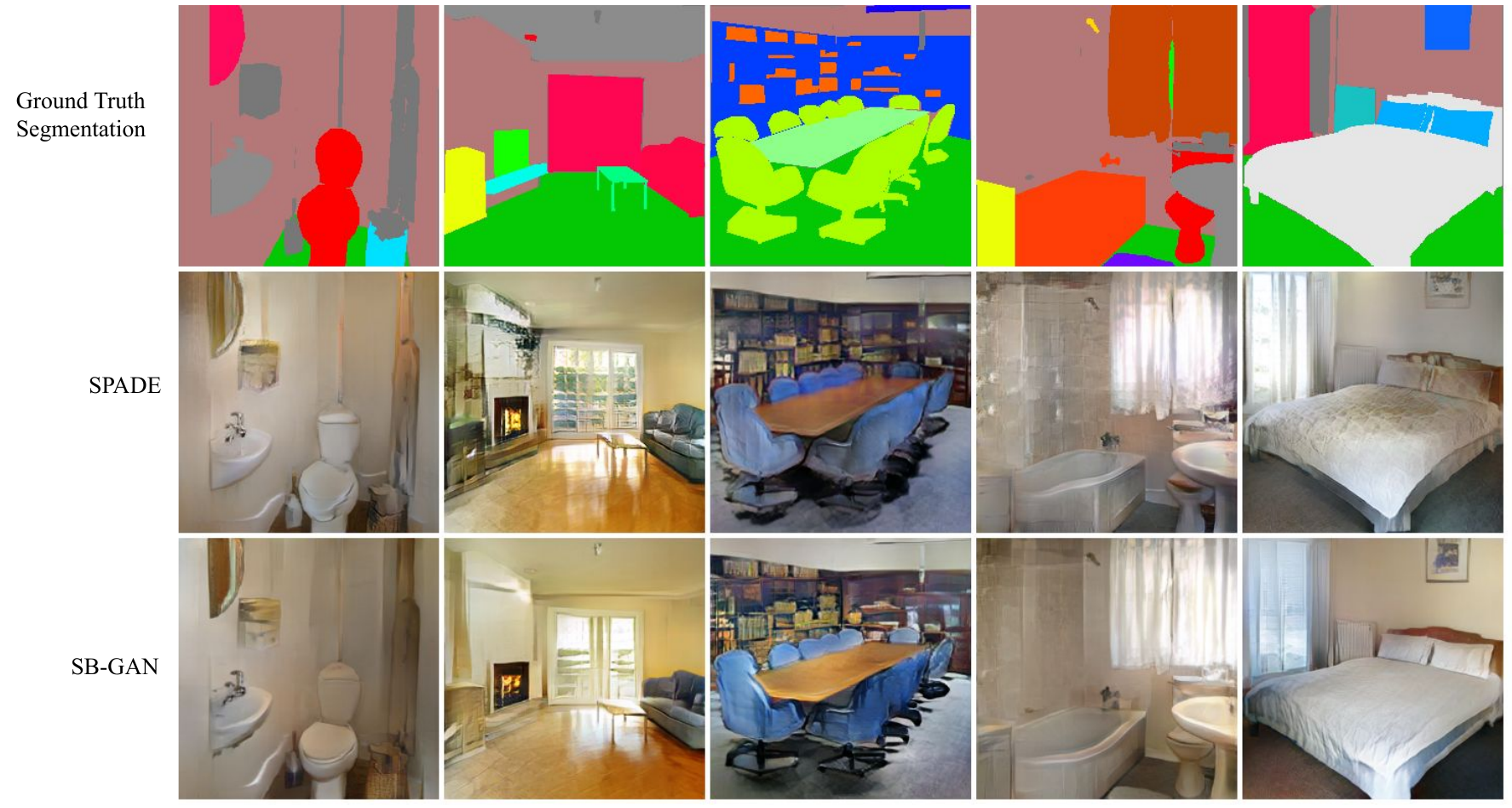}
\caption{The effect of SB-GAN on improving the performance of the state-of-the-art semantic image synthesis model (SPADE) on ground truth segmentations of ADE-Indoor validation set. For SB-GAN, we train the entire model end-to-end, extract the trained SPADE sub-network, and synthesize samples conditioned on the ground truth labels.}
\label{fig:ablate_ade_spade}
\end{figure*}

Moreover, Figures~\ref{fig:ablate_city_spade} and~\ref{fig:ablate_ade_spade} present additional examples illustrating the impact of SB-GAN on improving the performance of SPADE~\cite{SPADE}, the state-of-the-art semantic image synthesis model on ground truth segmentations. The third row in these two figures show examples of the synthesized images conditioned on ground truth labels when the SPADE sub-network is extracted from a trained SB-GAN model.

\begin{figure*}[t!]
\centering
\includegraphics[width=\textwidth]{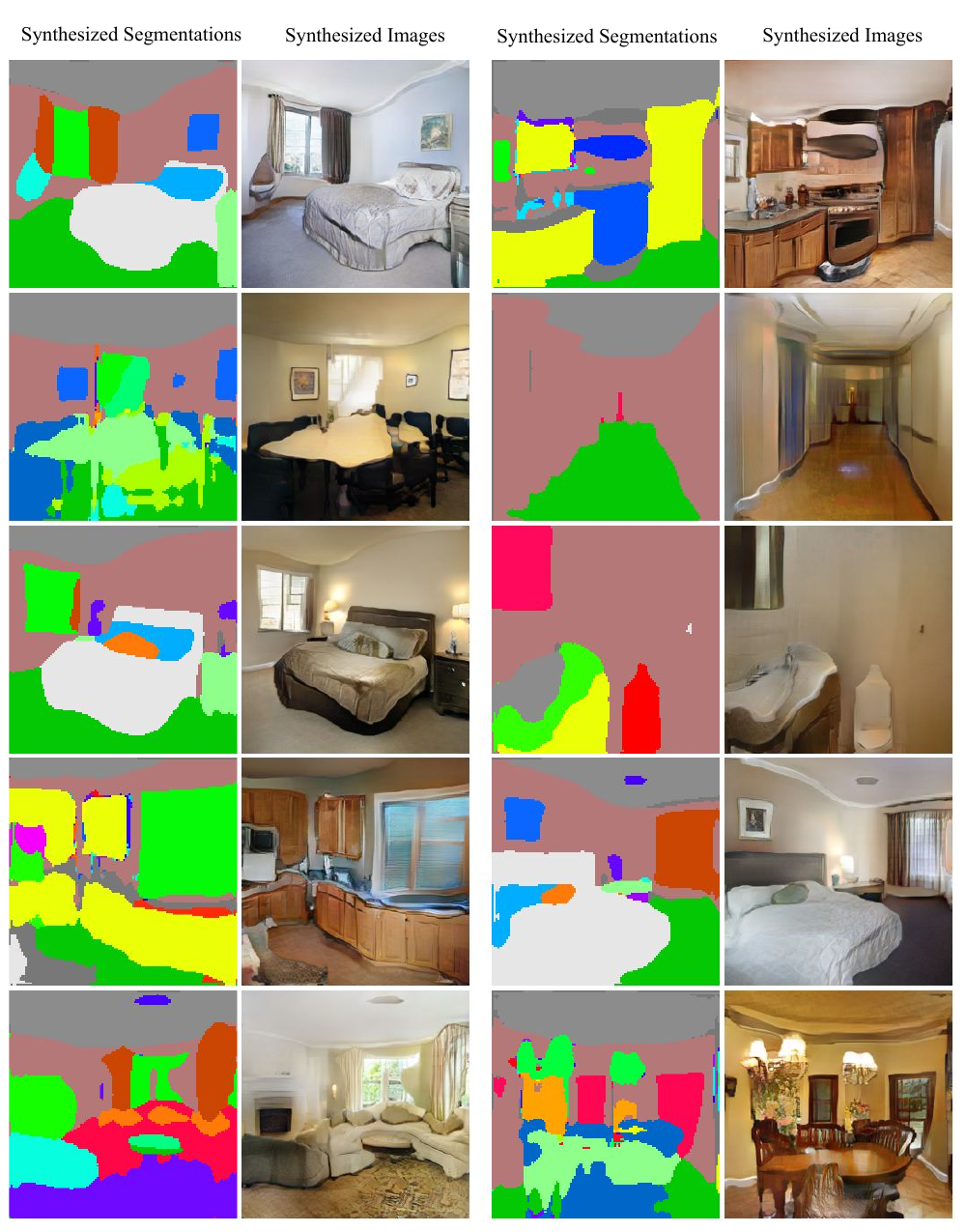}
\caption{Segmentations and their corresponding images synthesized by SB-GAN trained on the ADE-Indoor dataset.}
\label{fig:ablate_ade}
\end{figure*}

\end{document}

%% file: math_commands.tex
\usepackage{amsmath,amsfonts,bm}

\DeclareMathOperator{\FID}{FID} % fid

\def\eqref#1{equation~\ref{#1}}
\def\1{\bm{1}}

\DeclareMathAlphabet{\mathsfit}{\encodingdefault}{\sfdefault}{m}{sl}
\SetMathAlphabet{\mathsfit}{bold}{\encodingdefault}{\sfdefault}{bx}{n}

\DeclareMathOperator*{\argmax}{arg\,max}

\DeclareMathOperator{\Tr}{Tr}